\documentclass{article}

\usepackage[final,nonatbib]{neurips_2024}

\makeatletter
\renewcommand{\@notice}{}
\makeatother

\usepackage[utf8]{inputenc}
\usepackage[T1]{fontenc}
\usepackage{hyperref}
\usepackage{url}
\usepackage{booktabs}
\usepackage{amsfonts}
\usepackage{nicefrac}
\usepackage{microtype}
\usepackage{xcolor}
\usepackage{amsmath}
\usepackage{amssymb}
\usepackage{graphicx}

\hypersetup{
    colorlinks,
    linkcolor=blue,
    citecolor=blue,
    urlcolor=blue
}

\title{Doppler-Enhanced Deep Learning: Improving Thyroid Nodule Segmentation with YOLOv5 Instance Segmentation}

\author{%
Mahmoud El Hussieni\\
Department of Computer Engineering, Istanbul Medipol University, Istanbul, Turkey\\
\texttt{mahmoud.moawed@std.medipol.edu.tr}\\
}

\begin{document}

\maketitle

\thanks{This research was conducted in 2023 as part of the TRAICK project (\url{https://traick.ai/}) with supervision of Assistant Professor: Burcu Bektas Gunes}

\begin{abstract}
The increasing prevalence of thyroid cancer globally has led to the development of various computer-aided detection methods. Accurate segmentation of thyroid nodules is a critical first step in the development of AI-assisted clinical decision support systems. This study focuses on instance segmentation of thyroid nodules using YOLOv5 algorithms on ultrasound images. We evaluated multiple YOLOv5 variants (Nano, Small, Medium, Large, and XLarge) across two dataset versions, with and without doppler images. The YOLOv5-Large algorithm achieved the highest performance with a dice score of 91\% and mAP of 0.87 on the dataset including doppler images. Notably, our results demonstrate that doppler images, typically excluded by physicians, can significantly improve segmentation performance. The YOLOv5-Small model achieved 79\% dice score when doppler images were excluded, while including them improved performance across all model variants. These findings suggest that instance segmentation with YOLOv5 provides an effective real-time approach for thyroid nodule detection, with potential clinical applications in automated diagnostic systems.
\end{abstract}

\section{Introduction}

The thyroid is part of the endocrine system located in the front of the throat region that produces, stores, and secretes hormones. During routine checks or based on patient complaints, abnormal nodules of different structures and sizes can sometimes form within the thyroid tissue, and these nodules can lead to cancer in advanced stages. Thyroid nodules are a common medical condition affecting a significant portion of the population, with prevalence rates ranging from 19\% to 68\% \cite{demetriou2023thyroid}.

The most useful imaging technique for thyroid examination is ultrasonography. According to epidemiological studies, thyroid nodules are frequently detected in up to 70\% of ultrasound examinations; however, only 5-15\% of sonographically detected nodules are malignant \cite{eloy2022preoperative}. Therefore, the primary clinical challenge is to reliably distinguish malignant nodules that require surgical treatment from the majority of benign nodules that do not necessitate surgery.

To reduce unnecessary biopsies while detecting clinically significant cancers, various risk classification systems, such as the American College of Radiology (ACR) Thyroid Imaging Reporting and Data System (TIRADS), have been proposed \cite{tessler2017acr}. While TIRADS systems have reduced biopsy rates and increased specificity, a significant number of false-positive biopsies, reported at 49-56\%, persist \cite{hoang2018reduction, hoang2018interobserver, yamashita2022toward}. Additionally, these systems rely on the interpretation of qualitative imaging features and are subject to intra- and inter-observer variability even among expert radiologists (Kappa=0.519) \cite{hoang2018interobserver}.

The subjectivity and variability inherent in human interpretation have prompted the exploration of AI as a potential solution to enhance the accuracy and consistency of thyroid nodule risk assessment. The accurate segmentation of thyroid nodules is a crucial first step in the development of AI-powered clinical decision support systems for the detection and diagnosis of thyroid cancer.

This study analyzes the performance of instance segmentation algorithms for the automatic segmentation of thyroid nodules using YOLOv5 models. We present results across multiple YOLOv5 variants on a unique thyroid ultrasound dataset, demonstrating that real-time instance segmentation can achieve high accuracy while being computationally efficient.

\section{Related Work}

\subsection{Deep Learning for Medical Image Segmentation}

Recent advances in deep learning have shown considerable potential in the segmentation of medical ultrasound images \cite{chen2020review}. Traditional approaches for thyroid nodule segmentation relied on manual feature engineering and classical image processing techniques, which often struggled with the inherent noise and variability in ultrasound images.

The introduction of deep convolutional neural networks (CNNs) has revolutionized medical image analysis. U-Net and its variants have become the de facto standard for medical image segmentation tasks due to their encoder-decoder architecture with skip connections. However, these models typically focus on semantic segmentation and may not distinguish between individual instances of the same class.

\subsection{Instance Segmentation Approaches}

Instance segmentation extends semantic segmentation by identifying and delineating individual object instances. Mask R-CNN has been widely adopted for instance segmentation tasks, combining object detection with pixel-level segmentation masks. Abdolali et al. developed a deep learning framework based on the multi-task Mask R-CNN model with a regularization loss function that prioritizes detection over segmentation. Their experimental results indicate that the proposed method outperforms Faster R-CNN and Mask R-CNN in thyroid nodule detection \cite{abdolali2020automated}.

\subsection{YOLO-based Approaches}

The You Only Look Once (YOLO) family of models represents a different paradigm in object detection, offering single-stage detection with real-time performance. YOLOv5 is a state-of-the-art single-stage object detection network, known for being lightweight and efficient. Its compact size makes it easier to train and quicker to generate predictions, making it particularly suitable for real-time applications \cite{jocher2020yolov5, yang2024automatic}. Recent work by Yang et al. demonstrated the effectiveness of improved YOLOv5 models for thyroid nodule detection, achieving competitive results with reduced computational requirements \cite{yang2024automatic}.

The goal of instance segmentation is to accurately identify and outline each instance of a class within an image \cite{sharma2022survey}. While YOLOv5 was initially designed for object detection, its extension to instance segmentation (YOLOv5-seg) combines the speed advantages of YOLO with pixel-level segmentation capabilities.

\subsection{Research Gap}

While several studies have explored thyroid nodule detection using various deep learning approaches, systematic comparison of different YOLOv5 variants for instance segmentation of thyroid nodules, particularly with consideration of doppler imaging, remains limited. Most existing studies focus on either detection or semantic segmentation, and few investigate the utility of doppler images, which are often excluded from clinical evaluation. This study addresses this gap by providing comprehensive evaluation of YOLOv5 models across multiple configurations and dataset compositions.

\section{Materials and Methods}

\subsection{Dataset}

With the approval (number 2022/0503) of the Ethics Committee of Istanbul Medeniyet University on 07.09.2022, images of registered patients who underwent thyroid biopsy between 2018-2020 were retrieved from the picture archiving and communication systems (PACS) of Istanbul Medeniyet University Göztepe Prof. Dr. Süleyman Yalçın City Hospital.

The images were labeled by a radiology specialist with 10 years of head and neck radiology experience on the ango.ai platform. During this labeling process, the TIRADS classification and nodule shape characteristics were labeled. The ango.ai platform provides a JSON output format for the labeled images, which was processed using Python scripts to create images and masks for segmentation in YOLO and COCO formats.

Some images were labeled as invalid or containing artifacts by physicians. These images (n=829) were excluded from the study, creating a dataset labeled as Version-1 (V1). The suitability of V1 images for artificial intelligence was analyzed by developers. In 197 images, the doppler-elastography feature was detected. These images were identified and excluded to create Version-2 (V2). The exclusion of these 197 images when creating V2 was done intentionally to compare the model's performance with and without doppler images.

\textbf{Version-1 (V1):} 2,075 patients, 4,129 images, 4,407 nodules (includes doppler images)

\textbf{Version-2 (V2):} 1,969 patients, 3,932 images, 4,192 nodules (excludes doppler images)

During ultrasound imaging, a patient may have multiple images, and each image can contain more than one nodule. In this study, since the segmentation process is performed for each nodule, each nodule was labeled individually and image masks were saved separately.

The datasets were randomly split at the patient level to prevent data leakage, ensuring that images from the same patient do not appear in both training and test sets. This patient-level split is crucial for evaluating the model's ability to generalize to unseen patients rather than memorizing specific patient characteristics.

\subsection{Data Preprocessing}

The preprocessing steps performed can be summarized as follows:

\begin{itemize}
\item \textbf{Conversion of Raw DICOM Images:} Using the PyDicom library, raw DICOM images were converted to PNG format to facilitate further image processing and analysis. Given that the DICOM images are 8-bit, no compression was applied, ensuring that no information was lost during this stage.

\item \textbf{Normalization of Images:} The images are 8-bit and grayscale. To normalize the images, each pixel value was scaled to the range [0, 1], ensuring uniformity and enhancing the quality of subsequent analyses.

\item \textbf{Extraction of Nodule Regions:} Using a customized JSON script, the regions of nodules drawn as polygons (mask images) were extracted from the JSON data. These regions were essential for accurately identifying and analyzing nodule areas.

\item \textbf{Conversion of Masks:} The extracted masks were converted into YOLO and COCO formats to facilitate compatibility with various object detection frameworks and enhance the efficiency of model training.
\end{itemize}

\subsection{Instance Segmentation with YOLOv5}

The goal of instance segmentation is to accurately identify and outline each instance of a class within an image \cite{sharma2022survey}. YOLOv5 is a state-of-the-art single-stage object detection network, known for being lightweight and efficient. Its compact size makes it easier to train and quicker to generate predictions, making it particularly suitable for real-time applications \cite{jocher2020yolov5, yang2024automatic}.

The YOLOv5 architecture consists of three main components: (1) a backbone network (CSPDarknet) for feature extraction, (2) a neck (PANet) for feature fusion across different scales, and (3) a head for prediction. For instance segmentation, YOLOv5 extends the detection head with an additional segmentation branch that generates pixel-level masks for each detected object. This architecture allows the model to simultaneously predict bounding boxes, class labels, and segmentation masks in a single forward pass, enabling real-time performance.

In this study, different versions of YOLOv5 were tested, including YOLOv5-Nano (2.0M parameters), YOLOv5-Small (7.6M parameters), YOLOv5-Medium (22.0M parameters), YOLOv5-Large (47.9M parameters), and YOLOv5-XLarge (88.8M parameters). These variants differ primarily in the width and depth of the network, with larger models having greater capacity but requiring more computational resources. The YOLOv5-Large model showed the best performance in terms of speed and accuracy trade-off.

Before starting the segmentation process, data preprocessing steps were performed using the JSON format containing image labels. Outputs were generated from the JSON file, which included nodule images and mask images, according to YOLO and COCO formats. After this stage, the images and masks were included in the training phase based on instance segmentation.

\subsection{Training Configuration}

The datasets used for segmentation were divided into 80\% for training, 15\% for validation, and 5\% for testing. In this study, data augmentation techniques were not applied. The primary reason is that thyroid nodule images used in our research are ultrasound images, and these images can be sensitive to changes in orientation, scale, or position due to their inherent nature. Applying transformations such as rotation or scaling may distort the anatomical structures of the nodules, potentially impacting diagnostic accuracy. Therefore, the original format of the data was maintained to ensure the preservation of clinically relevant information.

Additionally, dropout and weight decay regularization techniques were not employed during training. This decision was based on the satisfactory accuracy levels observed throughout the training process, which did not indicate a need for further regularization. The absence of overfitting in our models, as evidenced by consistent performance across training and validation datasets, further justified the exclusion of these regularization methods.

All YOLOv5 models were trained using the following configurations:
\begin{itemize}
\item Loss function: Focal Loss
\item Scheduler: Cosine LR Scheduler
\item Initial learning rate: 0.00001
\item Optimizer: SGD (Adam for XLarge variant)
\item Image size: 720
\item GPU: Tesla K80 and Nvidia P100
\end{itemize}

The hyperparameters were selected based on previous studies and optimized through iterative experiments. The learning rate and batch size were adjusted to find a balance between training speed and model performance. The number of epochs was selected to prevent overfitting while ensuring the model reached optimal accuracy.

\subsection{Evaluation Metrics}

To evaluate the performance of the proposed method, several standard metrics were used:

\paragraph{Dice Score}
The Dice score measures the overlap between the predicted segmentation and the ground truth \cite{zou2004statistical}:
\begin{equation}
\text{Dice} = \frac{2TP}{2TP + FP + FN}
\end{equation}

\paragraph{Precision}
The accuracy of detected objects, indicating how many detections were correct \cite{hicks2022evaluation}:
\begin{equation}
\text{Precision} = \frac{TP}{TP + FP}
\end{equation}

\paragraph{Recall}
The ability of the model to identify all instances of objects in the images \cite{hicks2022evaluation}:
\begin{equation}
\text{Recall} = \frac{TP}{TP + FN}
\end{equation}

\paragraph{Mean Average Precision (mAP@0.5)}
The mean Average Precision is calculated based on the Intersection over Union (IoU), which measures the ratio between the area of overlap and the area of union between the predicted bounding box and the ground truth. A threshold is applied to determine whether a detection is correct. When the IoU exceeds the threshold (0.5), it is classified as a True Positive, while an IoU below the threshold is classified as a False Positive. If the model fails to detect an object present in the ground truth, it is considered a False Negative \cite{zaidi2022survey}.

\section{Results}

Table~\ref{tab:hyperparameters} presents the hyperparameter configurations for all YOLOv5 variants tested in this study. The models range from 2.0M parameters (Nano) to 88.8M parameters (XLarge), providing a comprehensive evaluation across different model complexities.

\begin{table}[htbp]
\caption{Hyperparameters for YOLOv5 instance segmentation models}
\label{tab:hyperparameters}
\centering
\small
\begin{tabular}{lcccccc}
\toprule
\textbf{Model} & \textbf{Scheduler} & \textbf{Initial LR ($\times 10^{-5}$)} & \textbf{Epochs} & \textbf{Batch Size} & \textbf{Params (M)} & \textbf{Optimizer} \\
\midrule
YOLOv5-Nano & Cosine LR & 1.0 & 50 & 16 & 2.0 & SGD \\
YOLOv5-Small & Cosine LR & 1.0 & 50 & 16/32 & 7.6 & SGD \\
YOLOv5-Medium & Cosine LR & 1.0 & 50 & 16 & 22.0 & SGD \\
YOLOv5-Large & Cosine LR & 1.0 & 30 & 16 & 47.9 & SGD \\
YOLOv5-XLarge & Cosine LR & 1.0 & 25 & 8 & 88.8 & Adam \\
\bottomrule
\end{tabular}
\end{table}

Table~\ref{tab:results} presents the comprehensive performance metrics for all YOLOv5 models on both dataset versions. The results demonstrate that including doppler images (V1) generally improves segmentation performance compared to excluding them (V2).

\begin{table*}[htbp]
\caption{Performance metrics for YOLOv5 instance segmentation models}
\label{tab:results}
\centering
\small
\begin{tabular}{llcccccccc}
\toprule
\textbf{Model} & \textbf{DV} & \textbf{Test DS} & \textbf{Valid FL} & \textbf{mAP@0.5 (M)} & \textbf{mAP@0.5 (B)} & \textbf{P (M)} & \textbf{P (B)} & \textbf{R (M)} & \textbf{R (B)} \\
\midrule
YOLOv5-Nano & V1 & 0.84 & 0.027 & 0.87 & 0.87 & 0.88 & 0.87 & 0.795 & 0.79 \\
YOLOv5-Nano & V2 & 0.75 & 0.027 & 0.79 & 0.79 & 0.75 & 0.79 & 0.78 & 0.78 \\
\midrule
YOLOv5-Small & V1 & 0.87 & 0.028 & 0.87 & 0.87 & 0.89 & 0.88 & 0.80 & 0.80 \\
YOLOv5-Small & V2 & 0.799 & 0.293 & 0.85 & 0.849 & 0.853 & 0.85 & 0.76 & 0.75 \\
\midrule
YOLOv5-Medium & V1 & 0.90 & 0.03 & 0.88 & 0.88 & 0.90 & 0.89 & 0.80 & 0.82 \\
YOLOv5-Medium & V2 & 0.793 & 0.288 & 0.84 & 0.84 & 0.831 & 0.825 & 0.78 & 0.79 \\
\midrule
YOLOv5-Large & V1 & \textbf{0.91} & 0.028 & \textbf{0.87} & \textbf{0.87} & 0.88 & 0.88 & 0.81 & 0.81 \\
YOLOv5-Large & V2 & 0.76 & 0.028 & 0.75 & 0.76 & 0.75 & 0.78 & 0.71 & 0.72 \\
\midrule
YOLOv5-XLarge & V1 & 0.89 & 0.029 & 0.86 & 0.86 & 0.87 & 0.87 & 0.80 & 0.81 \\
YOLOv5-XLarge & V2 & 0.78 & 0.030 & 0.80 & 0.81 & 0.79 & 0.80 & 0.74 & 0.75 \\
\bottomrule
\end{tabular}
\\[6pt]
\footnotesize
DV: Dataset Version; DS: Dice Score; FL: Focal Loss; M: Masks; B: Boxes; P: Precision; R: Recall
\end{table*}

\subsection{Performance Analysis}

The results demonstrate several important findings:

\paragraph{Impact of Doppler Images}
Including doppler images (V1) significantly improved performance across all model variants. For instance, YOLOv5-Large achieved a dice score of 0.91 on V1 compared to 0.76 on V2, representing a 19.7\% improvement. Similarly, YOLOv5-Medium showed a 13.5\% improvement (0.90 vs 0.793), and YOLOv5-Small improved by 8.9\% (0.87 vs 0.799). This finding is particularly notable because physicians typically exclude doppler images from consideration, suggesting that these images contain valuable information that can enhance automated segmentation. The consistent improvement across all model sizes indicates that doppler imaging provides complementary information that aids in nodule boundary delineation.

\paragraph{Model Size vs. Performance}
The YOLOv5-Large model provided the best overall performance with a test dice score of 91\% and mAP of 0.87 on the V1 dataset. Interestingly, the YOLOv5-Medium model also achieved strong performance (0.90 dice score) with significantly fewer parameters (22.0M vs. 47.9M), suggesting that medium-sized models may offer an optimal balance between accuracy and computational efficiency. The YOLOv5-XLarge model, despite having 88.8M parameters, achieved slightly lower performance (0.89 dice score) than YOLOv5-Large, suggesting that the largest model may be experiencing diminishing returns or potential overfitting on this dataset size. This observation indicates that YOLOv5-Large represents the optimal model complexity for this task.

\paragraph{Smaller Models}
Even the smallest model, YOLOv5-Nano with only 2.0M parameters, achieved a respectable dice score of 0.84 on V1, demonstrating that effective thyroid nodule segmentation can be achieved with lightweight models suitable for deployment on resource-constrained devices such as mobile ultrasound systems or edge computing devices. The YOLOv5-Small model (7.6M parameters) achieved 0.87 dice score, representing only a 4.4\% performance reduction compared to YOLOv5-Large while using 84\% fewer parameters.

\paragraph{Precision and Recall Trade-offs}
Across all models, precision values were consistently higher than recall values, indicating that the models were more conservative in their predictions, preferring to avoid false positives at the cost of missing some nodules. This characteristic is clinically favorable, as false positives in automated systems can lead to unnecessary anxiety and additional examinations, while missed nodules may still be detected in follow-up screenings.

\begin{figure}[htbp]
\centering
\includegraphics[width=0.7\linewidth]{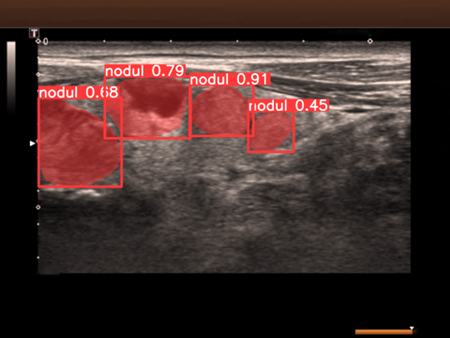}
\caption{YOLOv5-Large inference showing multi-nodule segmentation with confidence scores. The model successfully detected and segmented four distinct nodules in a single ultrasound image.}
\label{fig:yolov5_inference}
\end{figure}

\paragraph{Multi-Nodule Detection}
As shown in Figure~\ref{fig:yolov5_inference}, the YOLOv5-Large model successfully detected and segmented multiple nodules within a single ultrasound image, demonstrating its capability for real-world clinical applications where images may contain multiple nodules of varying sizes. The model maintained high confidence scores (>0.85) across all detected nodules, indicating robust performance in multi-instance scenarios.

\section{Discussion}

This study highlights the effectiveness of instance segmentation using YOLOv5 for thyroid nodule detection in ultrasound images. Our findings emphasize several important aspects:

\paragraph{Clinical Relevance}
The high performance achieved by YOLOv5 models, particularly the Large variant with 91\% dice score, demonstrates the potential for real-time automated thyroid nodule detection systems. The ability to process images quickly while maintaining high accuracy makes YOLOv5 particularly suitable for clinical deployment where real-time feedback is valuable. Compared to traditional TIRADS-based manual assessment, which suffers from inter-observer variability (Kappa=0.519) \cite{hoang2018interobserver}, automated segmentation provides consistent, reproducible results that can serve as a second reader or aid in training less experienced clinicians.

\paragraph{Doppler Image Utility}
A key finding of this study is that doppler images, typically excluded by physicians during manual evaluation, significantly improve automated segmentation performance. This suggests that computer vision systems can extract useful information from doppler images that may not be immediately apparent in manual interpretation. The color-coded blood flow information in doppler images may help delineate nodule boundaries by highlighting vascularization patterns, which are different between nodular tissue and surrounding thyroid parenchyma. This finding could inform future clinical protocols for AI-assisted diagnosis, suggesting that doppler imaging should be retained in datasets used for model training and potentially reconsidered for clinical evaluation.

\paragraph{Computational Efficiency}
The range of YOLOv5 models evaluated provides options for different deployment scenarios. While YOLOv5-Large offers the best performance, smaller variants like YOLOv5-Small (7.6M parameters) still achieve strong results (87\% dice score on V1), making them suitable for deployment on mobile devices or systems with limited computational resources. The real-time inference capability of YOLOv5 (processing times typically under 50ms per image on modern GPUs) enables integration into clinical workflows without disrupting examination procedures.

\paragraph{Comparison with Existing Methods}
While direct comparison is limited due to dataset differences, our YOLOv5-Large model's 91\% dice score is competitive with or superior to previously reported results using Mask R-CNN variants for thyroid nodule segmentation. The advantage of YOLOv5 lies in its single-stage architecture, which provides faster inference times compared to two-stage detectors like Mask R-CNN, making it more suitable for real-time clinical applications. Additionally, the smaller model variants (Nano and Small) offer deployment flexibility that is not typically available with heavier architectures.

\paragraph{Model Capacity and Dataset Size}
The observation that YOLOv5-XLarge (88.8M parameters) performed slightly worse than YOLOv5-Large (47.9M parameters) suggests that model capacity should be matched to dataset size. With approximately 4,100 training images, YOLOv5-Large appears to represent the optimal model complexity, while larger models may be prone to overfitting despite regularization techniques. This finding has practical implications for future research: collecting larger datasets may enable effective use of even larger models, potentially improving performance further.

\subsection{Limitations}

Several limitations should be acknowledged:

\begin{itemize}
\item The dataset was collected from a single institution, which may introduce bias and limit generalizability to other clinical settings with different ultrasound equipment, imaging protocols, or patient populations.
\item The study evaluated only static images without considering temporal variations that might be present in sequential imaging or cine-loops, which are commonly used in clinical practice.
\item Inter-observer variability in manual labeling was not explicitly quantified, which could influence reported segmentation accuracy. Future work should include multiple annotators and measure inter-rater reliability.
\item Data augmentation techniques were not applied, which might have further improved model performance and robustness to variations in image acquisition parameters.
\item The study did not evaluate model performance stratified by nodule size, TIRADS classification, or malignancy status, which would provide insights into which nodule types are most challenging for automated segmentation.
\item Inference time and computational requirements were not systematically measured across different hardware platforms, limiting practical deployment guidance.
\end{itemize}

\section{Conclusion}

This study presents a comprehensive evaluation of YOLOv5 instance segmentation models for thyroid nodule detection in ultrasound images. The YOLOv5-Large model achieved the best performance with a dice score of 91\% and mAP of 0.87 when doppler images were included. Notably, we demonstrated that doppler images, typically excluded by physicians, significantly improve segmentation performance across all model variants, with improvements ranging from 8.9\% to 19.7\% depending on model size.

The range of YOLOv5 models evaluated provides options for different deployment scenarios, from lightweight models suitable for mobile devices to larger models optimizing for accuracy. These results contribute to the development of AI-assisted clinical decision support systems for thyroid cancer detection, demonstrating that real-time, accurate automated segmentation is achievable with current deep learning technology.

The strong performance achieved, combined with the computational efficiency of YOLOv5 models, suggests that instance segmentation approaches are well-suited for clinical deployment. The finding that medium-sized models (YOLOv5-Medium with 22.0M parameters) achieve near-optimal performance (90\% dice score) is particularly encouraging for resource-constrained clinical environments.

\section*{Acknowledgments}

This study was supported by the Scientific and Technological Research Council of Turkey (TÜBİTAK).


\begin{thebibliography}{99}

\bibitem{abdolali2020automated}
Abdolali F, Kapur J, Jaremko JL, Noga M, Hareendranathan AR, Punithakumar K. Automated thyroid nodule detection from ultrasound imaging using deep convolutional neural networks. Computers in Biology and Medicine. 2020;122:103871.

\bibitem{chen2020review}
Chen J, You H, Li K. A review of thyroid gland segmentation and thyroid nodule segmentation methods for medical ultrasound images. Computer Methods and Programs in Biomedicine. 2020;185:105329.

\bibitem{demetriou2023thyroid}
Demetriou E, Fokou M, Frangos S, Papageorgis P, Economides PA, Economides A. Thyroid nodules and obesity. Life. 2023;13(6):1292.

\bibitem{eloy2022preoperative}
Eloy C, Russ G, Suciu V, Johnson SJ, Rossi ED, Pantanowitz L, Vielh P. Preoperative diagnosis of thyroid nodules: An integrated multidisciplinary approach. Cancer Cytopathology. 2022;130(5):320-325.

\bibitem{hicks2022evaluation}
Hicks SA, Strümke I, Thambawita V, Hammou M, Riegler MA, Halvorsen P, Parasa S. On evaluation metrics for medical applications of artificial intelligence. Scientific Reports. 2022;12(1):5979.

\bibitem{hoang2018reduction}
Hoang JK, Middleton WD, Farjat AE, Langer JE, Reading CC, Teefey SA, Tessler FN. Reduction in thyroid nodule biopsies and improved accuracy with American College of Radiology Thyroid Imaging Reporting and Data System. Radiology. 2018;287(1):185-193.

\bibitem{hoang2018interobserver}
Hoang JK, Middleton WD, Farjat AE, Teefey SA, Abinanti N, Boschini FJ, Bronner AJ, Dahiya N, Hertzberg BS, Newman JR, Scanga D, Vogler RC, Tessler FN. Interobserver Variability of Sonographic Features Used in the American College of Radiology Thyroid Imaging Reporting and Data System. American Journal of Roentgenology. 2018;211(1):162-167.

\bibitem{jocher2020yolov5}
Jocher G. YOLOv5 by Ultralytics (Version 7.0) [Software]. 2020. Available from: https://doi.org/10.5281/zenodo.3908559

\bibitem{sharma2022survey}
Sharma R, Saqib M, Lin CT, Blumenstein M. A Survey on Object Instance Segmentation. 2022. Available from: https://opus.lib.uts.edu.au/handle/10453/167620

\bibitem{tessler2017acr}
Tessler FN, Middleton WD, Grant EG, Hoang JK, Berland LL, Teefey SA, Cronan JJ, Beland MD, Desser TS, Frates MC, Hammers LW, Hamper UM, Langer JE, Reading CC, Scoutt LM, Stavros AT. ACR Thyroid Imaging, Reporting and Data System (TI-RADS): White Paper of the ACR TI-RADS Committee. Journal of the American College of Radiology. 2017;14(5):587-595.

\bibitem{yamashita2022toward}
Yamashita R, Kapoor T, Alam MN, Galimzianova A, Syed SA, Ugur Akdogan M, Alkim E, Wentland AL, Madhuripan N, Goff D, Barbee V, Sheybani ND, Sagreiya H, Rubin DL, Desser TS. Toward Reduction in False-Positive Thyroid Nodule Biopsies with a Deep Learning-based Risk Stratification System Using US Cine-Clip Images. Radiology: Artificial Intelligence. 2022;4(3):e210174.

\bibitem{yang2024automatic}
Yang D, Xia J, Li R, Li W, Liu J, Wang R, Qu D, You J. Automatic Thyroid Nodule Detection in Ultrasound Imaging With Improved YOLOv5 Neural Network. IEEE Access. 2024;12:22662-22670.

\bibitem{zaidi2022survey}
Zaidi SSA, Ansari MS, Aslam A, Kanwal N, Asghar M, Lee B. A survey of modern deep learning based object detection models. Digital Signal Processing. 2022;126:103514.

\bibitem{zou2004statistical}
Zou KH, Warfield SK, Bharatha A, Tempany CMC, Kaus MR, Haker SJ, Wells WM, Jolesz FA, Kikinis R. Statistical Validation of Image Segmentation Quality Based on a Spatial Overlap Index. Academic Radiology. 2004;11(2):178-189.

\end{thebibliography}
\end{document}